\documentclass[10pt,letterpaper,twocolumn]{article}
\usepackage[latin9]{luainputenc}
\usepackage{multirow}
\usepackage{amsmath}
\usepackage{amssymb}
\usepackage{graphicx}
\usepackage[unicode=true,
 bookmarks=false,
 breaklinks=true,pdfborder={0 0 1},backref=section,colorlinks=false]
 {hyperref}

\makeatletter

\providecommand{\tabularnewline}{\\}

\usepackage{cvpr}
\usepackage{times}
\usepackage{epsfig}

\cvprfinalcopy % *** Uncomment this line for the final submission

\def\cvprPaperID{****} % *** Enter the CVPR Paper ID here

\makeatother

\begin{document}
%%%%%%%%% TITLE

%\title{\LaTeX\ Author Guidelines for CVPR Proceedings}
%1. Efficient Encoding of Fine-tuned Neural Networks for Image Artifact Removal
%2. Weight-update Compression for Image Artifact Removal Neural Networks
%3. Efficient Fine-tuning of Decoder-side Neural Networks for Image Compression
%4. Joint Fine-tuning and Weight-update Compression for Image Artifacts Removal
%5. Compressing Weight-updates for Decoder-side Neural Networks
%5. Compressing Weight-updates for Decoder-side Artifacts Removal Neural Networks
%6. Compressing Weight-updates for Image Artifacts Removal Neural Networks
\title{Compressing Weight-updates for Image Artifacts Removal Neural Networks}
\author{Yat Hong LAM, Alireza Zare, Caglar Aytekin, Francesco Cricri,\\
 Jani Lainema, Emre Aksu, Miska Hannuksela\\
 Nokia Technologies\\
 Hatanpaan Valtatie 30, Tampere, Finland\\
 \texttt{\small{}{}{}{}yat.lam@nokia.com}{\small{}{}{}{} }}
\maketitle
\begin{abstract}
In this paper, we present a novel approach for fine-tuning a decoder-side
neural network in the context of image compression, such that the
weight-updates are better compressible. At encoder side, we fine-tune
a pre-trained artifact removal network on target data by using a compression
objective applied on the weight-update. In particular, the compression
objective encourages weight-updates which are sparse and closer to
quantized values. This way, the final weight-update can be compressed
more efficiently by pruning and quantization, and can be included
into the encoded bitstream together with the image bitstream of a
traditional codec. We show that this approach achieves reconstruction
quality which is on-par or slightly superior to a traditional codec,
at comparable bitrates. To our knowledge, this is the first attempt
to combine image compression and neural network's weight update compression.
\end{abstract}
%%%%%%%%% BODY TEXT

\section{Introduction}

There are two major directions in image compression: lossless compression
(e.g., PNG) and lossy compression (e.g., JPEG). In order to achieve
a high compression ratio and smaller file size, lossy compression
is widely applied in different areas including image storage and transmission.
Lossy compression methods usually introduce compression artifacts
into the decoded image, which greatly affect the perceptual quality
of the image. Some of the common compression artifacts are blocking
and quantization artifacts.

To alleviate the severeness of the problem, specific filters can be
used to remove the artifacts. Convolutional neural networks (CNNs)
have been used recently either within the traditional codec (e.g.
replacing some traditional filters as in \cite{Jia2019}) or after
it (e.g. a post-processing filter) \cite{ChaoDong2015}, \cite{LukasCavigelli2017}.
In \cite{yu2016deep}, the authors propose AR-CNN which aims to suppress
compression artifacts, Their network structure includes a skip connection
to bypass the network's layers.

In this paper, we present a novel approach for using a post-processing
neural network at decoder-side for artifacts removal, in the context
of image compression. We first pre-train the neural network filter
on a training dataset. At encoding time, we first encode and decode
the target image using a traditional codec. Then, the pre-trained
model is fine-tuned using the original and decoded target image, by
using an additional loss term which encourages the weight-update to
be sparse and close to quantized values. The final weight-update is
compressed by pruning and quantization, and included into the encoded
bitstream.

The work presented in this paper was used to participate to the 2019
Challenge on Learned Image Compression (CLIC). In particular, our
submission names were NTCodec2019vJ2 and NTCodec2019F4.

\section{Related Works}

Adaptive loop filtering (ALF) is a technique which was explored for
HEVC video compression standard \cite{chen2012adaptive}, where a
decoder-side filter is adapted to the input content by including into
the encoded bitstream all the filter coefficients. Instead, in our
case, the filter is a neural network and we adapt it by including
only a weight-update into the encoded bitstream. In \cite{tung2017fine},
the authors propose a method for jointly fine-tuning and compressing
a pre-trained neural network in order to adapt the network to a more
specialized domain than the pre-training domain, in order to avoid
overfitting due to over-parametrization. The authors compress the
whole network, whereas we compress only the weight-update, which is
more likely to require a low bitrate. In addition, they obtain a
compressed network which is different from the original pre-trained
network. Thus, the fine-tuned weights cannot be used to update a predefined
network structure. For the same reason, it is impossible to reset
the model to the pre-trained weights without storing a copy of the
pre-trained network.

This paper proposes a loss term which encourages compressibility of
the weight-update by achieving sparsity and more quantizable values.
In \cite{Aytekin2} and \cite{Aytekin2019mpeg}, the methodology of
compressive loss term was developed and applied to compress neural
networks' weights. We make use of a similar approach but apply it
on the weight-update of neural networks.

\section{Methodology}

Our proposed solution consists of using a traditional codec in combination
with a post-processing neural network which is applied on the full-resolution
decoded image. We chose the test model of Versatile Video Coding (VVC)
standard as the traditional codec, which is currently under development
\cite{Chen2019}.

The neural network filter is pre-trained in an offline phase. The
training images are first encoded and decoded using the VVC test model,
the decoded images, which are affected by compression artifacts, are
used as the input to the neural network, whereas the original uncompressed
images are used as the ground-truth. The neural network is trained
to remove the artifacts and reconstruct images with better visual
quality. As we aim to optimize the peak signal-to-noise ratio (PSNR)
of the filtered images, we use the mean squared error (MSE) as the
training loss, defined as $L_{mse}(I,\hat{I})=\frac{1}{N}\sum_{i}^{N}(I(i)-\hat{I}(i))^{2}$,
%\begin{equation}
%L_{mse}(I,\hat{I})=\frac{1}{N}\sum_{i}^{N}(I(i)-\hat{I}(i))^{2}\label{eq:mse}
%\end{equation}
where $I$ and $\hat{I}$ are the original and the reconstructed image,
respectively. The pre-trained weights then become part of the decoder
system.

\begin{figure*}
\hfill{}\includegraphics[width=1.5\columnwidth]{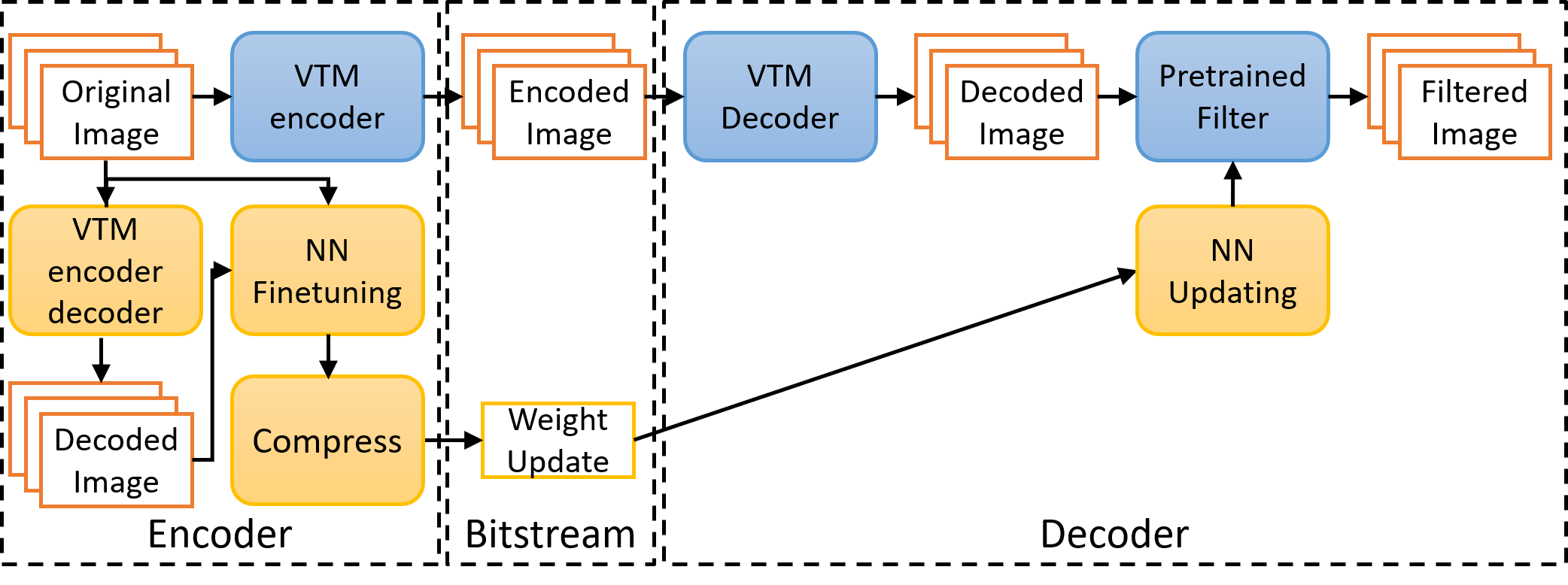}\hfill{}

\caption{\label{fig:Overview}Overview of the encoder-decoder structure. The
traditional path is shown in blue blocks and our proposed method is
shown in yellow blocks.}
\end{figure*}

In the online stage, i.e., during the encoding process, we further
fine-tune our network at encoder side on one or more VVC-decoded images,
as shown in Fig. \ref{fig:Overview}. Thus, the network is optimized
for the current test images. This fine-tuning process is performed
by using an additional training loss which encourages weight-updates
which are sparse and close to quantized values. The obtained weight-update
is then compressed and included into the bitstream, together with
the encoded images bitstream.

At decoder side, the weight-update is first decompressed and then
applied to the pre-trained neural network. The updated network is
applied on the VVC-decoded image for removing compression artifacts.
The network's weights are then reset to their original pre-trained
values, in order to be ready to be updated again for a new set of
images.

There are mainly two novel aspects in this work. Instead of including
the whole fine-tuned neural network into the bitstream, only the weight-update
of the network is included. To further reduce the bitrate, the weight-update
is made more compressible during training and is subsequently compressed
by pruning and non-uniform quantization.

\subsection{Network Structure}

The network structure is inspired by the U-net \cite{ronneberger2015u},
which consists of a contracting path and an expansive path. The contracting
path is formed by blocks consisting of a strided convolution layer,
a batch normalization layer and a leaky rectified linear unit (ReLU).
The number of feature filter is doubled at each block. Similarly the
expansive path is formed by blocks consisting of a transpose convolutional
layer, batch normalization layer and leaky ReLU activation. There are
lateral skip connections between the input of each block in the contracting
path and the output of each corresponding block in the expansive part.
A lateral skip connection is merged to the output of an expansive
block by concatenation.

We selected a fully-convolutional network because it requires less
parameters, which allows for using less data during training and fine-tuning,
and it also reduces the size of the weight-update. In addition, this
structure allows for handling input images of varying resolution.

\subsection{Fine-tuning with Weight-update Compression}

In the more common approach of having a pre-trained post-processing

network, the model is not adaptive to the content and it needs to generalize
to many types of content. To address this limitation, one approach
is to fine-tune or adapt the network on the test data. However, in
the context of networks used at decoder-side, including all the weights
into the bitstream requires a too high bitrate. This can be mitigated
by either using a small model, or by encoding a small weight-update
for a bigger network. This paper proposes a solution for the latter
approach, which has the advantage that for some test images the encoder
may decide not to send any weight-update and the decoder can use the
pre-trained network for filtering.

Fine-tuning can be done for each single test image, or for multiple
test images. In the first case we need to include into the bitstream
one weight-update per image. This approach can bring a higher
gain in visual quality, due to the homogeneity of the data,
but it is also more challenging to obtain a sufficiently small weight-update.
In the latter case we need to include a single weight-update for multiple
images, and the quality improvement depends on the
similarity among the images. We chose the latter approach.

The weight-update at each training iteration $t$ is commonly defined
as $\Delta w=-\rho\nabla_{w}L(w_{t})$, where $\rho$ is the learning
rate, $\nabla$ is the gradient operator, and $L$ is the loss function.
During fine-tuning, we are interested in the weight-update with respect
to the pre-trained network, thus we define the accumulated weight-update
at fine-tuning iteration $t$ as $\Delta w_{acc}=w_{t}-w_{0}$, where
$w_{0}$ are the pre-trained weights. Our goal is to compress this
accumulated weight-update. To this end, during fine-tuning we use
a combination of reconstruction loss and of weight-update compression
objective. The latter term aims to increase the compressibility of
the resultant weight-update, and is defined as follows:
\begin{equation}
L_{comp}(\Delta w_{acc})=\frac{|\Delta w_{acc}|}{||\Delta w_{acc}||}+\alpha\frac{||\Delta w_{acc}||^{2}}{|\Delta w_{acc}|}\label{eq:compLoss}
\end{equation}

The first term in Eq. \ref{eq:compLoss} is introduced in \cite{Hoyer1}
as a measure of the sparsity in a signal, and may also be referred

to as the sparsity term. It was also used in \cite{Aytekin2} for compressing a whole neural network. Minimizing this first term results in reducing
the number of non-zero values in the signal. The second term is added
to favour smaller absolute values for the non-zero weight-updates to regulate
the training and avoid exploding gradients. This second term was introduced in \cite{Aytekin2019mpeg} in the context of compressing a whole neural network. $\alpha$ is a regularizer
between the two terms.

The total loss function in the training is defined as a weighted sum
of MSE and the compression objective:
\begin{equation}
L_{total}=L_{mse}+\gamma L_{comp}\label{eq:totalLoss}
\end{equation}
where $\gamma$ is a dimensionless parameter to adjust the impact of the
compression objective.

\subsection{Post-training Compression}

Our hypothesis is that not all values in the weight-update are going
to have significant effect on the resultant image quality, and that
it is possible to preserve most of the improvement by discarding some
of the weight-update information.

The accumulated weight-update $\Delta w_{acc}$ is pruned by setting
values which are smaller than a threshold to zero, thus increasing
its sparsity. The remaining non-zero values are non-uniformly quantized
by $k$-means clustering. The compressed weight-update $\Delta w_{c}$
is then represented as follows: a flattened binary mask which indicates
the zeros and non-zeros in $\Delta w_{c}$, a tensor to store $k$-means
labels of the non-zero elements in $\Delta w_{c}$ and a dictionary
which maps the $k$-means labels to the corresponding value of cluster
centroids. These are all compressed into a single file by Numpy npz
algorithm.

The network is updated simply as $w_{rec}=w_{0}+\Delta w_{c}$, where
$w_{rec}$ are the reconstructed weights. The reconstructed image
$\hat{I}$ is obtained by filtering the VVC-decoded image with the
updated network.

\section{Experimental Results}

\begin{figure*}[!tp]
\hfill{}\includegraphics[width=1.2\columnwidth]{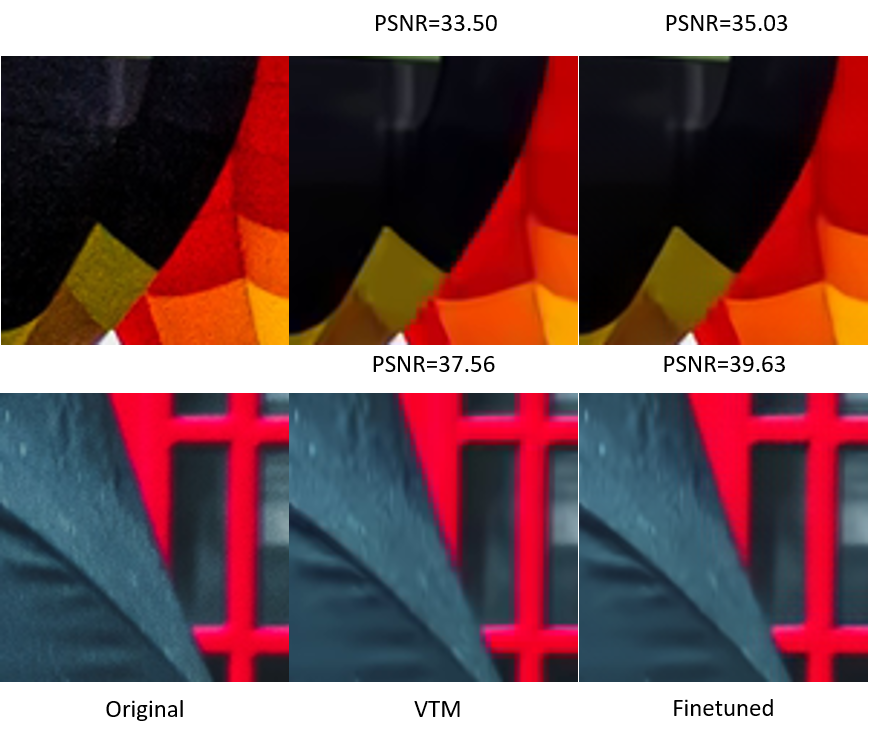}\hfill{}

\caption{\label{fig:Two-sample-patches}Two sample patches from CLIC test dataset
which show the improvement brought by our proposed finetuning (NTCodec2019vJ2)}
\end{figure*}

\subsection{Implementation Details}

We participated to the \textit{low-rate compression} track of the
CLIC competition, where the aim is to preserve the best image quality
with the limit of 0.15 bits-per-pixel (bpp). Although different image
quality metrics are considered in the competition, we focus on PSNR
as our image quality metric. The CLIC dataset was used for training
and testing. The training split contains 1632 images, whereas the
test split contains 330 images.

\subsubsection{Image Encoding}

The encoding was performed using the VVC Test Model VTM-4.0 with All
Intra (AI) configuration. The Test Model requires images in an uncompressed
raw format. Therefore, the source images in Portable Network Graphics
(PNG) format were converted to YUV 4:2:0 color sampling at 8 bits
per sample. The decoding operation is then followed by converting
the decoded images back to RGB. The quantization parameter (QP) was
chosen on a per-image basis to keep the bitrate of each image close
to 0.12 bpp for NTCodec2019F4 and 0.14 bpp for NTCodec2019vJ2. These
values allow for a margin that we use for including also a weight-update
into the bitstream, whose bitrate needs to be at most 0.03 bpp and
0.01 bpp, respectively.

\subsubsection{Neural Network Pre-training and Fine-tuning}

%During pre-training and fine-tuning, random patches of size $[255,255]$ are cropped from the original images and VVC-compressed images and used as the ground-truth and input data respectively.
The contracting part of our network has 3 blocks, in which convolutional
layers have stride 2, kernel-size 3x3, and numbers of channels 128,
256, 512. The expansive part has 3 blocks, in which the transpose
convolutional layers have stride 2, kernel-size 3x3, and numbers of
channels 256, 128, 3. Training and fine-tuning are performed using
Adam optimizer, and learning-rate $0.001$ for NTCodec2019F4 and $0.0005$
for NTCodec2019vJ2. To obtain further improvement in image quality,
we empirically found that it is better to fine-tune the network in
two phases. Initially, only MSE loss is used. After the MSE has dropped
to a certain level, the compression objective is added
to obtain a more compressible weight-update. The compression objective in Eq. \ref{eq:totalLoss} is weighted by $\gamma=m\frac{L_{mse}}{L_{comp}}$, where $m$ is an
empirically chosen hyper-parameter. No gradients are allowed to flow
through $\gamma$. The hyper-parameter $\alpha$ in Eq. \ref{eq:compLoss}
is chosen such that $\alpha\frac{||\Delta w_{acc}||^{2}}{|\Delta w_{acc}|}=\frac{1}{3}\frac{|\Delta w_{acc}|}{||\Delta w_{acc}||}$.
The post-training compression is done offline with different threshold
values $\tau$ and number of $k$-means clusters, $k$. The best weight-update
is selected based on two criteria: fulfilling the bpp margin requirement
(0.03 for NTCodec2019F4 and 0.01 for NTCodec2019vJ2), and resulting into best possible PSNR. The best compression
parameters for NTCodec2019F4 and NTCodec2019vJ2 were $\{\tau=0.00001,k=64\}$ and $\{\tau=0.005,k=4\}$, respectively.

%Table \ref{tab:hyperparams} provides other hyper-parameters and settings for the two submissions NTCodec2019J2 and NTCodec2019F4.

%\begin{table}
%\begin{centering}
%\begin{tabular}{|c|l|l|}
%\hline
% & (a) & (b) \tabularnewline
%\hline
%\hline
%\multicolumn{1}{|>{\centering}p{3cm}|}{VTM-encoded images} & 0.14 & 0.12\tabularnewline
%\hline
%\multicolumn{1}{|>{\centering}p{3cm}|}{Allowed bpp for weight update} & 0.01 & 0.03\tabularnewline
%\hline
%\multicolumn{1}{|>{\centering}p{3cm}|}{Compression multiplier $\gamma$} & 10 & \tabularnewline
%\hline
%\multicolumn{1}{|>{\centering}p{3cm}|}{Threshold $\tau$} & 0.005 & \tabularnewline
%\hline
%\multicolumn{1}{|c|}{No. of clusters $k$} & 4 & \tabularnewline
%\hline
%\end{tabular}
%\par\end{centering}
%\caption{\label{tab:hyperparams}Hyper-parameter and settings of submissions NTCodec2019J2 (a) and NTCodec2019F4 (b)}
%\end{table}

\subsection{Results}

The experimental results for NTCodec2019F4 and NTCodec2019vJ2 are shown
in Table \ref{tab:Experimental-result-on}. Two different post-processing
filters are pre-trained by using images encoded by VTM to 0.12 bpp
(NTCodec2019F4) and 0.14 bpp (NTCodec2019vJ2). These two networks
were fine-tuned in order to get two weight-updates. For NTCodec2019F4,
the PSNR of VTM-decoded images is improved by the pre-trained filter
and by the fine-tuned network by 0.21 dB and 0.26 dB, respectively. For
NTCodec2019vJ2, the PSNR of VTM-decoded images is improved by the pre-trained
filter and by the fine-tuned network by 0.17 dB and 0.22 dB, respectively.
Two example patches are shown in Fig. \ref{fig:Two-sample-patches}.

\begin{table}
\begin{centering}
\begin{tabular}{|l|c|c|}
\hline
Method  & PSNR  & \multirow{1}{*}{bpp}\tabularnewline
\hline 
\hline
%VTM4 0.12 bpp & 28.06 & 0.111 \tabularnewline
VTM4 0.12 bpp  & 28.13 & 0.111\tabularnewline
%VTM4 0.12 bpp + Pretrained Filter & 28.24 & -- \tabularnewline
VTM4 0.12 bpp + Pretrained Filter  & 28.34 & 0.111\tabularnewline
%\emph{Finetuned Filter 0.12 bpp} & 28.57 & 0.023 \tabularnewline
VTM4 0.12 bpp + \emph{Finetuned Filter}  & 28.39 & 0.134\tabularnewline
\hline
VTM4 0.14 bpp  & 28.80 & 0.1396\tabularnewline
VTM4 0.14 bpp + Pretrained Filter  & 28.97 & 0.1396\tabularnewline
VTM4 0.14 bpp + \emph{Finetuned Filter}  & 29.03 & 0.14\tabularnewline
\hline
VTM4 0.15 bpp  & 28.98 & 0.149\tabularnewline
\hline
\end{tabular}
\par\end{centering}
\caption{\label{tab:Experimental-result-on}Experimental results on CLIC test
dataset}
\end{table}

%-------------------------------------------------------------------------

\section{Conclusions}

In this paper, we described the work we used for our participation
to the 2019 Challenge on Learned Image Compression (CLIC). We presented
a novel method where a post-processing neural network, to be used
on images decoded by a traditional codec, is first pre-trained in
an offline stage, and then fine-tuned at encoding stage. In particular,
we proposed to jointly fine-tune and compress the weight-update by
using a weight-update compression objective. The compression objective
encourages weight-updates to be sparse and have values close to quantized
values. We experimented with different settings: one setting where
the weight-update can take up to 0.03 bpp, and one where the weight-update
can take up to 0.01 bpp. We used VTM as our traditional codec. In
the experiments, we showed that in both settings we get competitive
results with respect to using only VTM, thus proving that the approach
of moving information from the encoded images bistream to a neural
network is a promising new direction.

\noindent \medskip{}

{\small{}{}{}{}  \bibliographystyle{ieee_fullname}
\bibliography{egbib}
 }{\small\par}

\end{document}